\def\year{2020}\relax
\documentclass[letterpaper]{article} % DO NOT CHANGE THIS
\usepackage{aaai20}  % DO NOT CHANGE THIS
\usepackage{times}  % DO NOT CHANGE THIS
\usepackage{helvet} % DO NOT CHANGE THIS
\usepackage{courier}  % DO NOT CHANGE THIS
\usepackage[hyphens]{url}  % DO NOT CHANGE THIS
\usepackage{graphicx} % DO NOT CHANGE THIS
\usepackage{amsfonts}
\usepackage{amsmath}
\usepackage{graphicx}  % DO NOT CHANGE THIS
\usepackage{xcolor}
\usepackage{soul}
\usepackage{multirow}
\usepackage{multicol}
\usepackage{xspace}
\renewcommand\footnotemark{}
\newcommand{\projectname}{\emph{PCONV}\xspace}

\title{PCONV: The Missing but Desirable Sparsity in DNN Weight Pruning for Real-time Execution on Mobile Devices}
\author{\selectfont Xiaolong Ma\text{$^\dagger$}\textsuperscript{\rm 1}, Fu-Ming Guo\text{$^\dagger$}\textsuperscript{\rm 1}\thanks{\hspace{-6.0mm}$^\dagger$These authors contributed equally.}, Wei Niu\textsuperscript{\rm 2}, Xue Lin\textsuperscript{\rm 1}, Jian Tang\textsuperscript{\rm 3,4},\\ {\bf \Large Kaisheng Ma\textsuperscript{\rm 5}, Bin Ren\textsuperscript{\rm 2}, Yanzhi Wang\textsuperscript{\rm 1}}\\ 
\textsuperscript{\rm 1}Northeastern University, 
\textsuperscript{\rm 2}College of William and Mary, 
\textsuperscript{\rm 3}DiDi AI Labs, 
\textsuperscript{\rm 4}Syracuse University, 
\textsuperscript{\rm 5}Tsinghua University \\
{E-mail:} \textsuperscript{\rm 1}\{ma.xiaol, guo.fu\}@husky.neu.edu, \textsuperscript{\rm 1}\{xue.lin, yanz.wang\}@northeastern.edu,\\
\textsuperscript{\rm 2}wniu@email.wm.edu, \textsuperscript{\rm 2}bren@cs.wm.edu, \textsuperscript{\rm 3}tangjian@didiglobal.com, \textsuperscript{\rm 5}kaisheng@mail.tsinghua.edu.cn
}
 
\begin{document}

\maketitle

\begin{abstract}
Model compression techniques on Deep Neural Network (DNN) have been widely acknowledged as an effective way to achieve acceleration on a variety of platforms, and DNN weight pruning is a straightforward and effective method. There are currently two mainstreams of pruning methods representing two extremes of pruning regularity: \emph{non-structured}, fine-grained pruning can achieve high sparsity and accuracy, but is not hardware friendly; \emph{structured}, coarse-grained pruning exploits hardware-efficient structures in pruning, but suffers from accuracy drop when the pruning rate is high. 
% \textcolor{red}{In this paper, we introduce \projectname, a new dimension, fine-grained structured pruning through employing pre-designed fine-grained pruning patterns on each convolution kernel. THIS SENTENCE HAS PROBLEM.} 
In this paper, we introduce \projectname, comprising a new sparsity dimension, -- fine-grained pruning patterns inside the coarse-grained structures. 
\projectname comprises two types of sparsities, Sparse Convolution Patterns (SCP) which is generated from intra-convolution kernel pruning and connectivity sparsity generated from inter-convolution kernel pruning. 
Essentially, SCP enhances accuracy due to its special vision properties, and connectivity sparsity increases pruning rate while maintaining balanced workload on filter computation. 
To deploy \projectname, we develop a novel compiler-assisted DNN inference framework and execute \projectname models in real-time without accuracy compromise, which cannot be achieved in prior work. Our experimental results show that, 
\projectname outperforms three state-of-art end-to-end DNN frameworks, TensorFlow-Lite, TVM, and Alibaba Mobile Neural Network with speedup up to $39.2\times$, $11.4\times$, and $6.3\times$, respectively, with no accuracy loss. 
% Real-time inference of representative large-scale DNNs (e.g., VGG-16, ResNet-50) can be achieved using mobile devices.
Mobile devices can achieve real-time inference on large-scale DNNs.

\end{abstract}

\section{Introduction}
Deep neural network (DNN) has emerged as the fundamental element and core enabler in machine learning applications due to its high accuracy, excellent scalability, and self-adaptiveness~\cite{goodfellow2016deep}. A well trained DNN model can be deployed as inference system for multiple objectives, such as image classification~\cite{krizhevsky2012imagenet}, object detection~\cite{ren2015faster}, and natural language processing~\cite{hinton2012deep}. However, the state-of-art DNN models such as VGG-16~\cite{simonyan2014very}, ResNet-50~\cite{he2016deep} and MobileNet~\cite{howard2017mobilenets} involve intensive computation and high memory storage, making it very challenging to execute inference system on current mobile platforms in a real-time manner.

% Recently, along with the rapid emergence of high-end mobile devices, executing DNNs on mobile platforms gains popularity and is quickly becoming the mainstream\textcolor{red}{[ref]} for broad applications such as sensor nodes, wireless access points, smartphones, wearable devices, video streaming, augmented reality, robotics, unmanned vehicles, smart health devices, etc.\textcolor{red}{[ref]}.
Recently, high-end mobile platforms are rapidly overtaking desktop and laptop as primary computing devices for broad DNN applications such as wearable devices, video streaming, unmanned vehicles, smart health devices, etc.~\cite{philipp2011sensor}\cite{lane2015early}\cite{boticki2010quiet}.
Developing a real-time DNN inference system is desirable but still yield to the limited computation resources of embedded processors on a mobile platform. Multiple end-to-end mobile DNN acceleration frameworks, such as TVM~\cite{chen2018tvm}, TensorFlow-Lite (TFLite)~\cite{TensorFlow-Lite} and Alibaba Mobile Neural Network (MNN)~\cite{Ali-MNN}, have been developed. However, the inference time of large-scale DNNs (e.g., 242ms inference time using TVM on Adreno 640 GPU with VGG-16) is still far from real-time requirement.

In order to mitigate the challenge brings by the DNN's bulky computation and achieve the goal of real-time inference, it is necessary to consider algorithm-level innovations. Various DNN model compression techniques are studied, among which \emph{weight pruning}~\cite{han2015deep}\cite{mao2017exploring}\cite{dai2017nest}\cite{wen2016learning}\cite{he2017channel} can result in a notable reduction in the model size. Early work~\cite{han2015deep} on \emph{non-structured} weight pruning (fine-grained) prunes weights at arbitrary location, resulting in a sparse model to be stored in the compressed sparse column (CSC) format. It leads to an undermined processing throughput because the indices in the compressed weight representation cause stall or complex workload on highly parallel architectures~\cite{han2015deep}\cite{wen2016learning}. On the other hand, \emph{structured} weight pruning~\cite{wen2016learning} (coarse-grained) is more hardware friendly. By exploiting filter pruning and channel pruning, the pruned model is more regular in its shape, which eliminates the storage requirement in weight indices. However, it is observed that structured pruning hurts accuracy more significantly than non-structured sparsity.

It is imperative to find a new granularity level that can satisfy high accuracy demand as well as regularity in DNN model structure. We make the observation that non-structured and structured pruning are two extremes of the full design space. The two missing keys are: (i) Find a new, intermediate sparsity dimension that can fully leverage both the high accuracy from fine-grained model and high regularity level from coarse-grained model; (ii) Find the corresponding (algorithm-compiler-hardware) optimization framework which can seamlessly bridge the gap between hardware efficiency and the new sparsity dimension. To address the above problems, this paper proposes \projectname, comprising (a) a new sparsity dimension that exploits both intra-convolution and inter-convolution kernel sparsities, exhibiting both high accuracy and regularity, and revealing a previously \emph{unknown} point in design space; and (b) a \emph{compiler-assisted DNN inference framework} that fully leverages the new sparsity dimension and achieves real-time DNN acceleration on mobile devices.

In \projectname, we call our intra-convolution kernel pruning \emph{pattern pruning} and inter-convolution kernel pruning \emph{connectivity pruning}. 
For pattern pruning, 
% we constrain the number of non-zeros in a kernel to be a constant, thus generate arbitrary shape for each kernel. 
a fixed number of weights are pruned in each convolution kernel. 
% This method seems to be similar to non-structured weight pruning, but the difference is that, the sparsity ratio in each filter is uniformly distributed and there is only a limited number of pattern shapes. 
Different from non-structured weight pruning, pattern pruning produces the same sparsity ratio in each filter and a limited number of pattern shapes. 
Essentially, our designed patterns correspond to the computer vision concept of key convolution filters, 
% for the purpose of smoothing and/or sharpening. 
such as Gaussian filter for smoothing, Laplacian of Gaussian filter for smoothing and sharpening. 
For connectivity pruning, the key insight is to {\em cut the connections} between certain input and output channels, which is equivalent to removal of corresponding kernels, making filter ``length" shorter than original model. 
With connectivity pruning, we further enlarge compression rate and provide greater DNN acceleration potential, while maintaining balanced workload in filter-wise computation of DNNs. Pattern and connectivity pruning can be combined at algorithm level and accelerated under the unified compiler-assisted acceleration framework.  
For our advanced \emph{compiler-assisted DNN inference framework}, we use execution code generation which converts DNN models into computational graphs and applies multiple optimizations including a high-level, fine-grained DNN layerwise information extraction, filter kernel reorder and load redundancy elimination. All design optimizations are general, and applicable to both mobile CPUs and GPUs.

We demonstrate that pattern pruning consistently improve model accuracy. When combined with connectivity pruning, the results still outperform current DNN pruning methods, both non-structured and structured weight pruning. 
In Section ``Accuracy Analysis", we show \projectname is the most desirable sparsity among current prune-for-acceleration works. 
We also deploy \projectname model on our compiler-assisted mobile acceleration framework and compare with three state-of-art frameworks on mobile CPU and GPU, TensorFlow Lite, TVM, and MNN, using three widely used DNNs, VGG-16, ResNet-50, and MobileNet-v2 and two benchmark datasets, ImageNet and CIFAR-10. 
Evaluation results show that \projectname achieves up to $39.2\times$ speedup without any accuracy drop. Using Adreno 640 embedded GPU, \projectname achieves an unprecedented 19.1 ms inference time of VGG-16 on ImageNet dataset. 
% To the authors' knowledge it is the first time to achieve real-time execution of such representative large-scale DNNs on mobile devices. 
To the best of our knowledge, it is the first time to achieve real-time execution of such representative large-scale DNNs on mobile devices.

\begin{figure}[t]
    \centering
    \includegraphics[width=0.44 \textwidth]{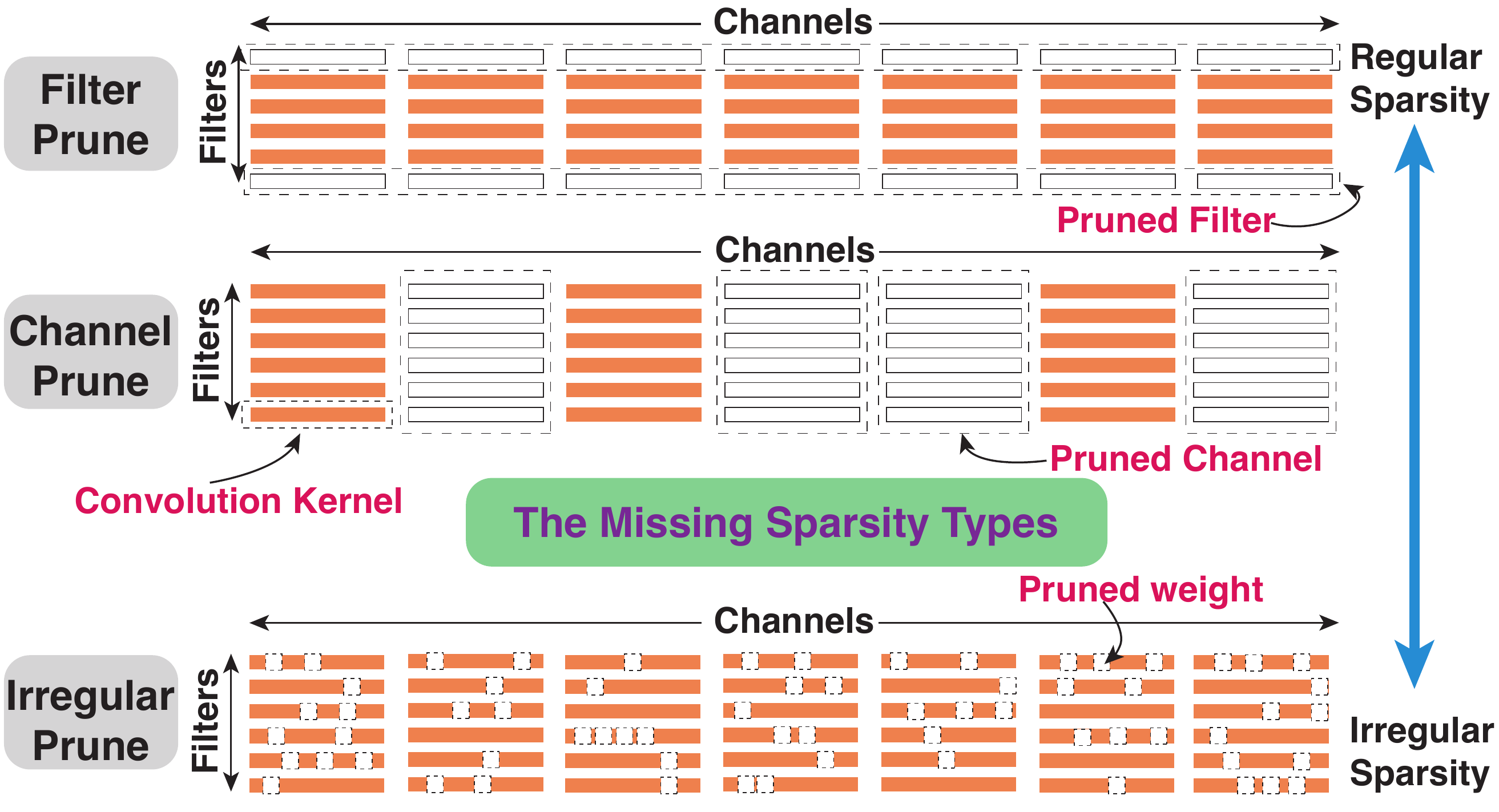}
    %\vspace{-1mm}
    \caption{Overview of different weight pruning dimensions.}
    \label{fig:sparsity_type}
    % \vspace{-3mm}
\end{figure}
\section{Background}
\subsection{DNN Model Compression}
DNN model compression is a promising method to remove redundancy in the original model. It targets on the purpose that inference time can be reduced if fewer weights are involved in the computation graph. The weight pruning method acts as a surgeon to remove the inherently redundant neurons or synapses. As Figure~\ref{fig:sparsity_type} shows, two main approaches of weight pruning are the general, non-structured pruning and structured pruning, which produce irregular and regular compressed DNN models, respectively. 

\textbf{Non-structured pruning:} 
Early work is~\cite{han2015deep}, in which an iterative, heuristic method is used with limited, non-uniform model compression rates. Flourished by~\cite{zhang2018systematic} and \cite{ren2019ADMMNN} with the powerful ADMM~\cite{boyd2011distributed} optimization framework, non-structured pruning achieves very high weight reduction rate and promising accuracy. However, for compiler and code optimization, irregular weight distribution within kernels requires heavy control-flow instructions, which degrades instruction-level parallelism. Also, kernels in different filters have divergent workloads, which burdens thread-level parallelism when filters are processed through multi-threading. Moreover, irregular memory access causes low memory performance and thereby execution overheads.

\textbf{Structured pruning:} 
This method has been proposed to address the index overhead and imbalanced workload caused by non-structured pruning. Pioneered by~\cite{wen2016learning}\cite{he2017channel}, structured weight pruning generates regular and smaller weight matrices, eliminating overhead of weight indices and achieving higher acceleration performance in CPU/GPU executions. However, it suffers from notable accuracy drop when the pruning rate increases.

\subsection{Patterns in Computer Vision}
Convolution operations exist in different research areas for an extended period of time, such as image processing, signal processing, probability theory, and computer vision. 
In this work, we focus on the relationship between conventional image processing and state-of-art convolutional neural networks in the usage of convolutions. In image processing, the convolution operator is manually crafted with prior knowledge from the particular characteristics of diverse patterns, such as Gaussian filter. On the other hand, in convolutional neural networks, the convolution kernels are randomly initialized, then trained on large datasets using gradient-based learning algorithms for value updating.

\cite{NIPS2014_5348} derived a network architecture named Convolutional Kernel Networks (CKN), with lower accuracy than current DNNs, thus limited usage. \cite{zhang2019shiftinvar} proposed to apply the blur filter to DNNs before pooling to maintain the shift-equivalence property. The limited prior work on the application of conventional vision filters to DNNs require network structure change and do not focus on weight pruning/acceleration, thus distinct from \projectname. 

\subsection{DNN Acceleration Frameworks on Mobile Platform}
Recently, researchers from academia and industry have investigated DNN inference acceleration frameworks on mobile platforms, including 
% DeepX~\cite{lane2016deepx}, 
TFLite~\cite{TensorFlow-Lite}, 
% DeepEar~\cite{lane2015deepear}, 
TVM~\cite{chen2018tvm}, 
Alibaba Mobile Neural Network (MNN)~\cite{Ali-MNN}, 
DeepCache~\cite{xu2018deepcache} 
% DeepMon~\cite{huynh2017deepmon}, 
and DeepSense~\cite{yao2017deepsense}. 
% and MCDNN~\cite{han2016mcdnn}. 
These works do not account for model compression techniques, and the performance is far from real-time requirement. There are other researches that exploit model sparsity to accelerate DNN inference, e.g., 
\cite{liu2015sparse}, 
% DeftNN~\cite{hill2017deftnn}, 
SCNN~\cite{parashar2017scnn}, 
% AdaDeep~\cite{liu2018demand}, 
but they either do not target mobile platforms (require new hardware) or trade off compression rate and accuracy, thus having different challenges than our work.

\section{Motivations}

Based on the current research progress on DNN model compression vs. acceleration, we analyze and rethink the whole design space, and are motivated by the following three points:

\textbf{Achieving both high model accuracy and pruning regularity.}
In non-structured pruning, any weight can be pruned. This kind of pruning has the largest flexibility, thus achieves high accuracy and high prune rate. But it is not hardware-friendly. On the other hand, structured pruning produces hardware-friendly models, but the pruning method lacks flexibility and suffers from accuracy drop. Our motivation is to use the best of the above two sparsities. To achieve that, we introduce a new dimension, pattern-based sparsity, revealing a previously unknown design point with high accuracy and structural regularity simultaneously.

\textbf{Image enhancement inspired sparse convolution patterns.}
The contemporary DNN weight pruning methods originate from the motivation that eliminating redundant information (weights) will not hurt accuracy. On the other hand, these pruning methods scarcely treat pruning as a specific kind of binary convolution operator, not to mention exploiting corresponding opportunities. Along this line, we find that sparse convolution patterns have the potential in enhancing image quality thanks to its special vision properties.
% with limitation of binary numerical representation, 
% Furthermore, we find that pattern-based convolution operators such as \textit{Gaussian} filter and \textit{Laplacian of Gaussian} (LoG) filter 
% would have genuine conventional computer vision meaning and enhance the image quality, . 
% we propose our pattern-based weight pruning method, ~\projectname, to utilize the mathematical vision meaning of diverse patterns, and pattern selection algorithm to distribute distinct patterns thoroughly balanced in the neural model.
% \textcolor{red}{Motivated by the fact that the sparse convolution pattern may increase DNNs' performance, thus provides greater possibility to include more sparsity type (in our case, the connectivity pruning) and larger compression rate, we need to design a set of sparse convolution patterns which benefits image classification performance in different DNN structures.} \textcolor{red}{DO YOU CALL THE ABOVE THING A SENTENCE? WHAT DO YOU WANT TO SAY??????}
Motivated by the fact that sparse convolution patterns can potentially enhance image quality, we propose our carefully designed patterns which are derived from mathematical vision theory.

\textbf{Compiler-assisted DNN inference framework.}
% \textcolor{red}{The proposed pattern-based sparsity preserves the high accuracy due to its fine-grained patterns, the question is how to re-gain similar hardware efficiency as coarse-gained structured pruning.} \textcolor{red}{DO YOU HAVE THE BASIC KNOWLEDGE OF GRAMMAR?} 
With the higher accuracy enabled by 
fine-grained pruning patterns, the key question is 
how to re-gain similar (or even surpass) hardware efficiency 
as coarse-gained structured pruning. 
We take a unique approach and design an optimized, compiler-assisted DNN inference framework to close the performance gap between full structured pruning and pattern-based pruning.

\begin{figure}[t]
    \centering
    \includegraphics[width=0.429\textwidth]{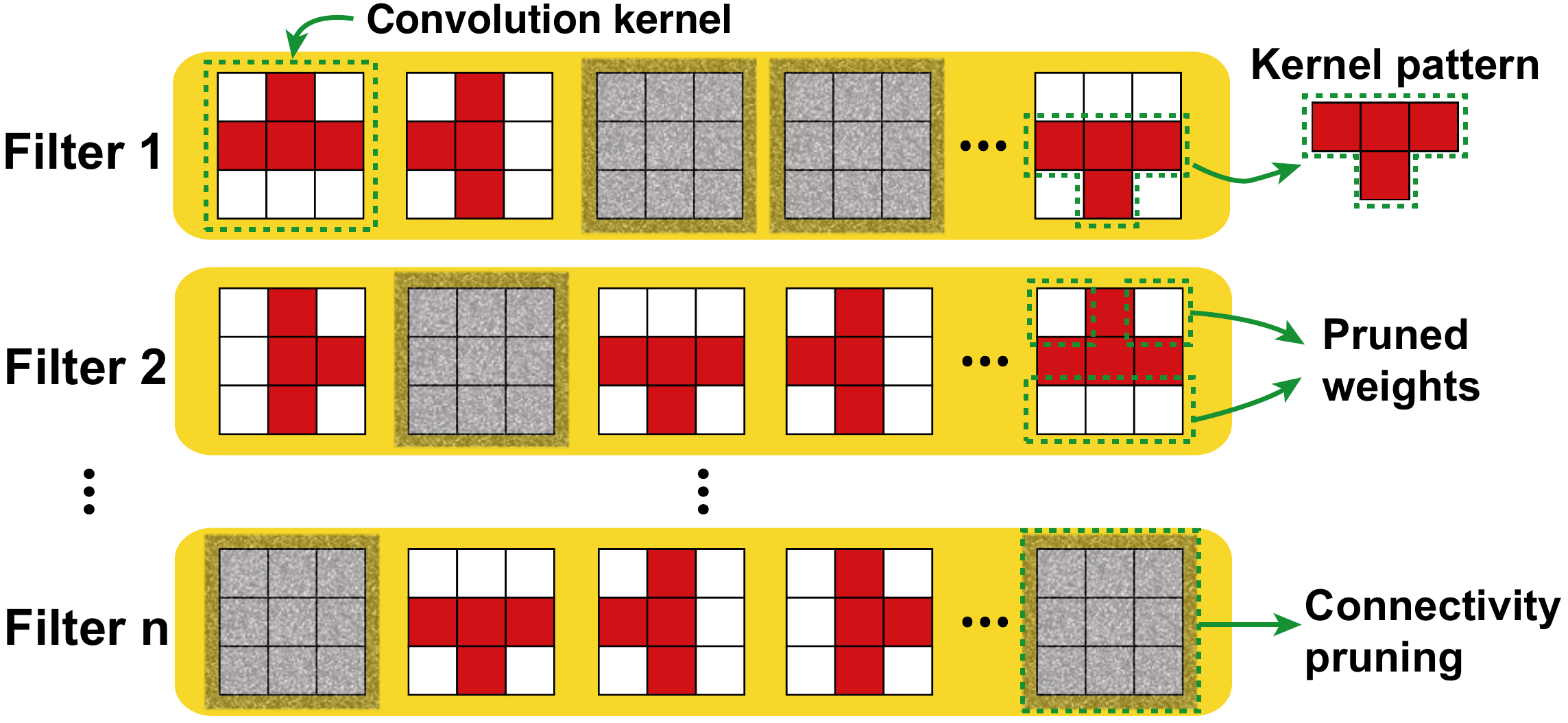}
    %\vspace{-1mm}
    \caption{Illustration of pattern pruning and connectivity pruning.}
    \label{fig:pattern_pruning}
    % \vspace{-3mm}
\end{figure}

\section{Theory of Sparse Convolution Patterns (SCP)}
% \textcolor{blue}{version1}

% Let an image with resolution $H \times W$ be represented by $X\in\mathbb{R}^{H \times W \times 3}$. 
% Let an $L-$layers DNN be represented as a feature extractor $\mathcal{F}(X) \in \mathbb{R}^{H \times W \times F \times C}$, comprising filter kernels $K$ with dimension ${H \times W}$, filters $F$ and channels $C$. To each layer $l\!\in\!\{1, \dots, L\}$, the corresponding filter kernels are $K_{l}$ with dimension ${H_{l} \times W_{l}}$.

% \textcolor{blue}{Let an image with resolution $H \times W$ be represented by $X \in \mathbb{R}^{H \times W \times 3}$. An $L-$layer DNN can be expressed as a feature extractor $ \mathcal{F}_{l}(\mathcal{F}_{l-1}(\ldots\mathcal{F}_{1}(X)\ldots)) $, with layer index $l \in\{1, \dots, L\}$. Inside the DNN, each convolutional layer is defined as $\mathcal{F}_{l}(X) \in \mathbb{R}^{H_{l} \times W_{l} \times F_{l} \times C_{l}}$, with filter kernel shape ${H_{l} \times W_{l}}$, filters $F_{l}$ and channels $C_{l}$.} 
% with shape size $H_{l} \times W_{l}$ and each element $p_{i}$ in pattern $P$ obeys to the value limitation $p_{i} \in \left\{ 0, 1 \right\}$. 
% All the possible S compose the SCP Space $P=\left\{P_{1},P_{2}, \ldots, P_{n}\right\}$.
Let an image with resolution $H \times W$ be represented by $X \in \mathbb{R}^{H \times W \times 3}$. An $L-$layer DNN can be expressed as a feature extractor $ \mathcal{F}_{L}(\mathcal{F}_{L-1}(\ldots\mathcal{F}_{1}(X)\ldots)) $, with layer index $l \in\{1, \dots, L\}$. Inside the DNN, each convolutional layer is defined as $\mathcal{F}_{l}(X_l) \in \mathbb{R}^{H_{l} \times W_{l} \times F_{l} \times C_{l}}$, with filter kernel shape ${H_{l} \times W_{l}}$, number of filters $F_{l}$ and number of channels $C_{l}$.

% \textcolor{green}{Compare the above two paras.}
% \textcolor{blue}{Got it! Thanks very much, Professor!}

Besides treating pruning as a redundant information removal technique, we consider it as incorporating an additional convolution kernel $P$ to perform element-wise multiplication with the original kernel.
$P$ is termed the \emph{Sparse Convolution Pattern} (SCP), 
% with shape size $H_{l} \times W_{l}$ and each element $p_{i}$ in pattern $P$ obeys to the value limitation $p_{i} \in \left\{ 0, 1 \right\}$. 
% All the possible S compose the SCP Space $P=\left\{P_{1},P_{2}, \ldots, P_{n}\right\}$. 
with dimension $H_{l} \times W_{l}$ and binary-valued elements (0 and 1). 
Specific SCPs fit the mathematical vision theory well according to our following derivation. Based on the mathematical rigority, we propose the novel \emph{pattern pruning} scheme, i.e., applying SCPs to convolution kernels. As illustrated in Figure~\ref{fig:pattern_pruning}, the white blocks denote a fixed number of pruned weights in each kernel. The remaining red blocks in each kernel have arbitrary weight values, while their locations form a specific SCP $P_{i}$. Different kernels can have different SCPs, but the total number of SCP types shall be limited.

In order to further increase the pruning ratio and DNN inference speed, we can selectively cut the connections between particular input and output channels, which is equivalent to the removal of corresponding kernels. This is termed \emph{connectivity pruning}. \emph{Connectivity pruning} is illustrated in Figure~\ref{fig:pattern_pruning}, with gray kernels as pruned ones. The rationale of connectivity pruning stems from the desirability of locality in layerwise computations inspired by human visual systems~\cite{yamins2016using}. It is a good supplement to pattern pruning. Both pruning schemes can be integrated in the same algorithm-level solution and compiler-assisted mobile acceleration framework.

\subsection{The Convolution Operator}

In conventional image processing, a convolution operator is formally defined by the following formula, where the output pixel value $g(x, y)$ is the weighted sum of input pixel values $f(x, y)$, and $h(k, l)$ is the weight kernel value
\begin{equation}
\fontsize{9}{8.5}\selectfont
    g(x, y)=\sum_{k, l} f(x+k, y+l) h(k, l)
\end{equation}
This formula could transform to
\begin{equation}
\fontsize{9}{8.5}\selectfont
    g(x, y)=\sum_{k, l} f(k, l) h(x-k, y-l)
    \label{eq:2}
\end{equation}
Then we derive the notation of convolution operator as:
\begin{equation}
\fontsize{9}{8.5}\selectfont
    g=f * h
\end{equation}

Convolution is a linear shift-invariant (LSI) operator, satisfying the commutative property, the superposition property and the shift-invariance property. 
% \textcolor{red}{Should be property other than principle.}
Additionally, convolution satisfies the associative property following the Fubini's theorem. 
% \textcolor{red}{WHAT is the meaning of associative principle? Associated? What do you want to say?}

% \st{Albeit the employment of SCP to the DNN model occurs after training} \textcolor{red}{HOW CAN THIS BE AFTER DNN TRAINING???}\st{, the commutative property and the shift-invariance property guarantee that the application of SCP still maintains the image pre-processing property.} \textcolor{red}{WHAT IS THE MEANING OF IMAGE PRE-PROCESSING PROPERTY? IS IT A PROPERTY? NO. UNCLEAR HERE.} 

\subsection{Sparse Convolution Pattern (SCP) Design}
Our designed SCPs could be transformed to a series of steerable filters~\cite{Freeman1991}, i.e., the Gaussian filter and Laplacian of Gaussian filter, which function as image smoothing, edge detection or image sharpening in mathematical vision theory. 

\textbf{Gaussian filter:}
Consider a two-dimensional Gaussian filter $G$:
\begin{equation}
\fontsize{9}{8.5}\selectfont
    G(x, y, \sigma)=\frac{1}{2 \pi \sigma^{2}} e^{-\frac{x^{2}+y^{2}}{2 \sigma^{2}}}
\end{equation}
$x$ and $y$ are input coordinates, and $\sigma$ is standard deviation of the Gaussian distribution.
Typically, the Gaussian filter performs image smoothing, and further sophisticated filters can be created by first smoothing the image input with a unit area Gaussian filter, then applying other steerable filters.
% \textcolor{red}{Again not clear here. Then do what after the first step?}

\textbf{Laplacian of Gaussian filter:}
The Laplacian operator is the second derivative operator.
According to the associative property, smoothing an image with Gaussian filter and then applying Laplacian operator is equivalent to convolve the image with the Laplacian of Gaussian (LoG) filter:
\begin{equation}
\fontsize{9}{8.5}\selectfont
    \nabla^{2} G(x, y, \sigma)=\left(\frac{x^{2}+y^{2}}{\sigma^{4}}-\frac{2}{\sigma^{2}}\right) G(x, y, \sigma)  
\end{equation}
The LoG filter is a bandpass filter that eliminates both the high-frequency and low-frequency noises. LoG has elegant mathematical properties, and is valid for a variety of applications including image enhancement, edge detection, and stereo matching.

\textbf{Taylor series expansion} is utilized to determine the approximate values of the LoG filter with $3 \times 3$ filter size. First, we consider the 1-D situation. The Taylor series expansions of 1-D Gaussian filter $G(x)$ are given by: 
\begin{small}
    \begin{equation}
        G(x\!+\!h)\!=\!G(x)\!+\!h G^{\prime}(x)\!+\!\frac{1}{2} h^{2} G^{\prime \prime}(x)\!+\!\frac{1}{3 !} h^{3} G^{\prime \prime \prime}(x)\!+\!O\left(h^{4}\right)\label{eq:taylor_1}
    \end{equation}
\end{small}
\begin{small}
    \begin{equation}
        G(x\!-\!h)\!=\!G(x)\!-\!h G^{\prime}(x)\!+\!\frac{1}{2} h^{2} G^{\prime \prime}(x)\!-\!\frac{1}{3 !} h^{3} G^{\prime \prime \prime}(x)\!+\!O\left(h^{4}\right)\label{eq:taylor_2}
    \end{equation}
\end{small}
By summing \eqref{eq:taylor_1} and \eqref{eq:taylor_2}, we have
\begin{equation}
\fontsize{9}{8.5}\selectfont
G(x+h)+G(x-h)=2 G(x)+h^{2} G^{\prime \prime}(x)+O\left(h^{4}\right)\label{eq:taylor_3}
\end{equation}
The second derivative of Gaussian $G^{\prime \prime}(x)$ is equivalent to LoG $\nabla^{2} G(x)$. Equation~\eqref{eq:taylor_3} is further transformed to 
\begin{equation}
\fontsize{9}{8.5}\selectfont
\frac{G(x-h)-2 G(x)+G(x+h)}{h^{2}}\!=\!\nabla^{2} G(x)\!+\!O\left(h^{2}\right)\label{eq:laplacian_ref}
\end{equation}
Applying central difference approximation of LoG $\nabla^{2} G(x)$, we derive the 1-D approximation of LoG filter as $\left[\begin{smallmatrix}{1} & {-2} & {1} \end{smallmatrix}\right]$. Then we procure the 2-D approximation of LoG filter by convolving $\left[\begin{smallmatrix}{1} & {-2} & {1} \end{smallmatrix}\right]$ and $\left[\begin{smallmatrix}{1} \\ {-2} \\ {1}\end{smallmatrix}\right]$, and get result as $\left[\begin{smallmatrix}{-1} & {2} & {-1} \\ {2} & {-4} & {2} \\ {-1} & {2} & {-1}\end{smallmatrix}\right]$. 
According to the property of second derivative:
\begin{equation}
\fontsize{9}{8.5}\selectfont
	\nabla^{2} G(x, y)=G_{x x}(x, y)+G_{y y}(x, y)
\end{equation}
and Equation \eqref{eq:laplacian_ref}, we have
\begin{equation}\label{eq:plus_filter}
\fontsize{9}{8.5}\selectfont
    G_{x x}(x, y)+G_{y y}(x, y)\!=\!\left(\left[\begin{smallmatrix}{1} & {-2} & {1}\end{smallmatrix}\right]\!+\!\left[\begin{smallmatrix}{1} \\ {-2} \\ {1}\end{smallmatrix}\right]\right) * G(x, y)
\end{equation}
Based on \eqref{eq:plus_filter}, we derive another approximation of LoG as $\left[\begin{smallmatrix}{0} & {1} & {0} \\ {1} & {-4} & {1} \\ {0} & {1} & {0}\end{smallmatrix}\right]$.

According to the central limit theorem, the convolution of two Gaussian functions is still a Gaussian function, and the new variance is the sum of the variances of the two original Gaussian functions. Hence, we convolve the above two approximations of LoG and then apply normalization, and get the \emph{Enhanced Laplacian of Gaussian} (ELoG) filter as 
$\left[\begin{smallmatrix}{0} & {1} & {0} \\ {1} & {8} & {1} \\ {0} & {1} & {0}\end{smallmatrix}\right]$.

% In the SCP space, due to the representation limitation of only ``0" and ``1", we use probability density estimation to make the further approximation.

\cite{powerofinterpolation} have proved the convergence of the interpolation in the context of (multi-layer) DNNs, so we utilize the interpolated probability density estimation to make the further approximation.
In ELoG filter
% $\left[\begin{smallmatrix}{0} & {1} & {0} \\ {1} & {8} & {1} \\ {0} & {1} & {0}\end{smallmatrix}\right]$
where 1 appears,
% we take an equal probability $(1-p)$ to mask an arbitrary position as 0 randomly. 
we mask it to 0 with the probability of $(1-p)$. 
Because we uniformly convolve SCPs into $n$ convolutional layers, this random masking operation can be treated as distributed interpolation of SCPs. 
In continuous probability space, interpolating SCPs into convolution function is a specific Probability Density Function (PDF), 
% and interpolating SCPs into the convolution function 
so the effect of interpolating SCPs is accumulating probability expectations of interpolation into $n$ convolutional layers. 
Besides, the convolution function is normalized to unity, so we separate the coefficient $p$ in the following equation.
% normalized such that the area under the function is equal to unity.
% \begin{equation}
%     \begin{split}
    
%     \left.\underbrace{\left[\begin{matrix}{0} & {1} & {0} \\ {1} & {1} & {1} \\ {0} & {0} & {0}\end{matrix}\right] \dots\left[\begin{matrix}{0} & {1} & {0} \\ {1} & {1} & {0} \\ {0} & {1} & {0}\end{matrix}\right] \dots\left[\begin{matrix}{0} & {0} & {0} \\ {1} & {1} & {1} \\ {0} & {1} & {0}\end{matrix}\right] \dots\left[\begin{matrix}{0} & {1} & {0} \\ {0} & {1} & {1} \\ {0} & {1} & {0}\end{matrix}\right]}_{\text{n times interpolation}}\right.\label{eq:prune_pattern}
    
%     \left.=\left[\begin{array}{lll}{0} & {p} & {0} \\ {p} & {1} & {p} \\ {0} & {p} & {0}\end{array}\right]^{n}=\left[p\left[\begin{array}{111}{0} & {1} & {0} \\ {1} & {1 / p} & {1} \\ {0} & {1} & {0}\end{array}\right]\right]^{n}\right. \hspace{26}
%     \end{split}
% \end{equation}
\begin{figure}[h!]
    \centering
    \includegraphics[width=0.43 \textwidth]{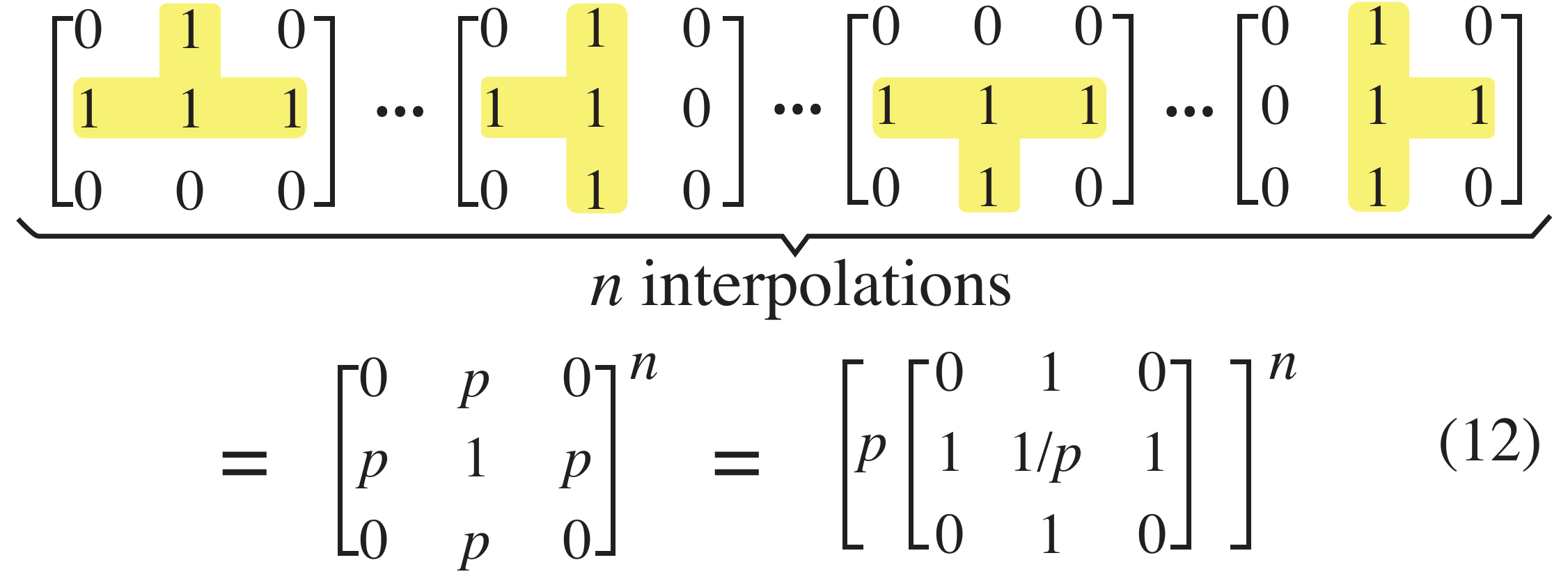}
    %\vspace{-1mm}
    % \caption{Overview of different weight pruning dimensions. \textcolor{red}{Will re-draw a new figure}}
    % \label{fig:sparsity_type}
    % \vspace{-3mm}
\end{figure}
The four SCPs are shown in colored positions in (12). In order to get the best approximation to 
% filter $\left[\begin{smallmatrix}{0} & {1} & {0} \\ {1} & {8} & {1} \\ {0} & {1} & {0}\end{smallmatrix}\right]$, 
ELoG filter, 
% which is the application of two \emph{LoG} filters by our formula derivation, 
we set $p=0.75$ and $n=8$, then the desired filter is equal to interpolating these four SCPs for eight times. The coefficient $p$ has no effect after normalization.

\textbf{Upper bound:} According to ~\cite{blakemore1969}, the optimal times for applying the \textit{LoG} 
% \textcolor{green}{LoG or ELoG?} 
% \textcolor{blue}{LoG, the ELoG is self-defined and for ELoG the number would be 3 and 5} 
filter is six and the maximum is ten. Thus the desired number of times to interpolate 
% \textcolor{red}{interpolated apply in grammarly wrong. Maybe interpolate SCP.} 
the SCP in (12) is around 24 and the maximum number is around 55. This upper bound covers most of the existing effective DNNs, even for ResNet-152, which comprises 50 convolutional layers with filter kernel size of $3\times3$.

The four SCPs in (12) form the ELoG filter through interpolation. Hence, the designed SCPs inherit the de-noising and sharpening characteristics of LoG filters. We visualize the intermediate results of DNNs to interpret and verify the advancement of our designed SCPs in the following section.
% \textcolor{red}{Continue working on this para?} 
% \textcolor{blue}{OK}

\subsection{Visualization and Interpretation}

% \textcolor{red}{To Xiaolong: Checked the section?} \textcolor{purple}{ -Yes, already checked.}
% The current DNN for vision tasks can be explained as a self-learned or supervised-learned deep feature extractor, in which individual channel captures different features. Then the learning system makes the prediction based on these features. 
% The main content of this para moved to the above joining part.
% According to our formula derivation, appropriately applying our designed SCPs is equivalent to the application of a series of LoG filters. Hence, the designed SCPs inherit the de-noising and sharpening characteristics of LoG filters. In this section, we visualize the intermediate results of DNN to interpret and verify the advancement of our designed SCPs.
% as interpolated convolution filters will achieve enhanced DNN performance,
% as the blurring, de-noising and sharpening characteristics of SCP help the extraction of features more effectively. 
% of the learning system could be enhanced along with the model size compressed. 
% As SCP provides a more visional meaningful regularization to approach the final loss of the system, 
% the blurring, de-noising and sharpening 
% and anti-alias 
% meanings of SCP help to extract features more effectively in distinct channels of the DNN.

\begin{figure}[t]
    \centering
    \includegraphics[width=0.47 \textwidth]{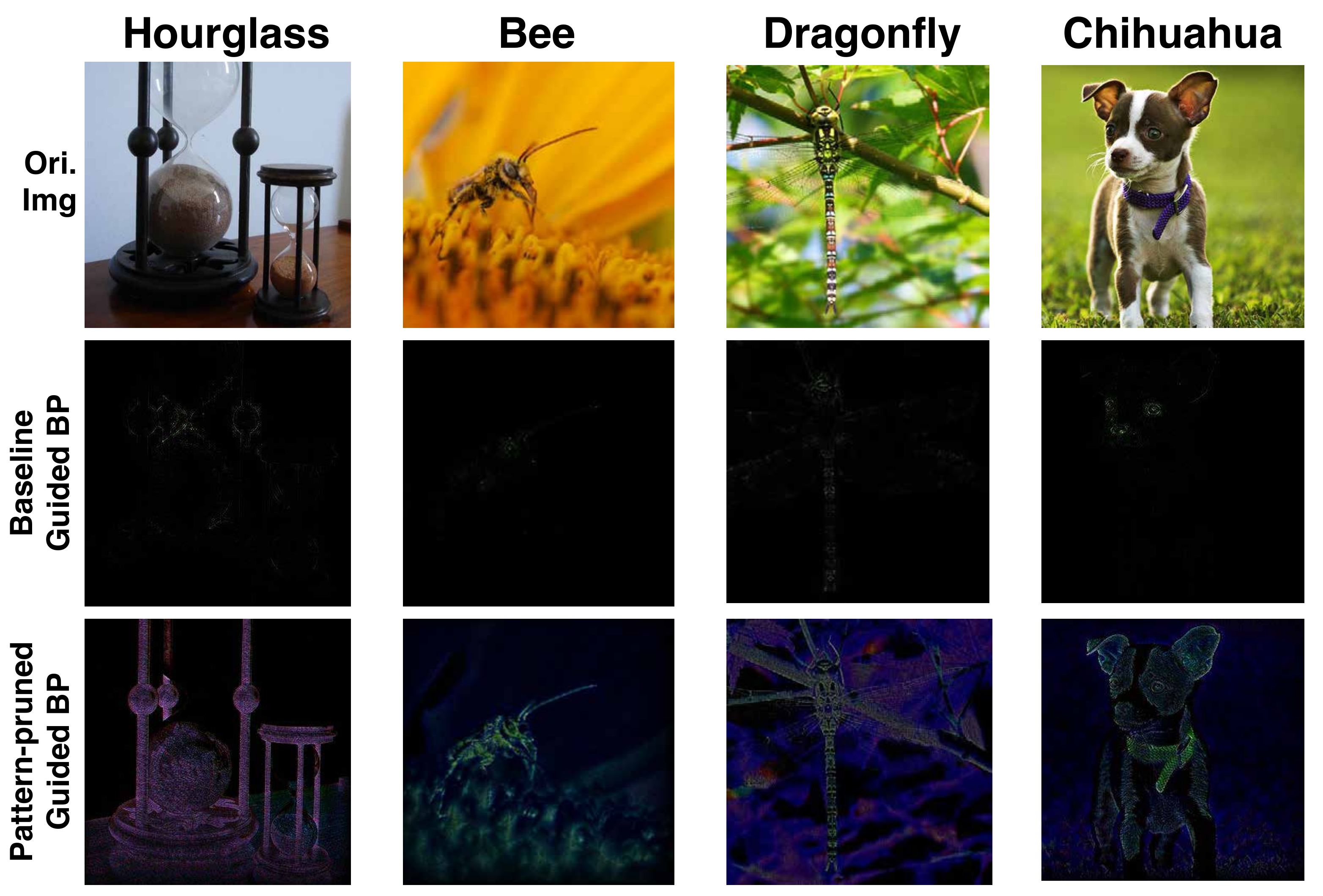}
    %\vspace{-1mm}
    \caption{Visualization of intermediate results (\emph{saliency map} of \emph{gradient images}) in original VGG-16 model and pattern pruned VGG-16 model through \emph{guided-backpropagation}.}
    \label{fig:visualization}
    % \vspace{-3mm}
\end{figure}
Explanations of individual DNN decision have been explored by generating informative heatmaps such as CAM and grad-CAM~\cite{gradcam}, or through guided-backpropagation (BP)~\cite{guided-backpropagation} conditioned on the final prediction. Utilizing guided-backpropagation, we can visualize what a DNN has learned. 
% including the original VGG-16 baseline model and the pruned VGG-16 model applied with SCP, which indeed improves the information quality of the intermediate results of the DNN by our visualization verification. 
The visualization results of applying SCPs to an original DNN model (\emph{pattern pruning}) are demonstrated in Figure~\ref{fig:visualization}. We sample four input images from the ImageNet dataset, as ``hourglass", ``bee", ``dragonfly" and ``chihuahua", then apply the guided-backpropagation to propagate back from each target class label 
and get the gradient images. 
% \textcolor{red}{towards? What does it mean here?} 
% \textcolor{blue}{(Guided-BP does not calculate the real gradients, instead calculate another kind of gradients from the real label, will rewrite here)}
Eventually, we generate the \emph{saliency maps} of gradient images. Compared with the original VGG-16 model, the pattern pruned VGG-16 model captures more detailed information of the input image with less noise. 

% There are plenty of DNN visualization techniques. In Supplemental Materials, we demonstrate two more sets of visualization results using integrated gradients and inverted representation methods. And both sets show our pattern pruned model collects more information in an image than original model. 
We conclude that by applying our designed SCPs, \emph{pattern pruning} enhances DNNs' image processing ability, which will potentially enhance the inference accuracy of a DNN.
% and the most direct effect is reflected on DNNs' accuracy performance. We will analyze the accuracy results in the following section.

\begin{figure}[t]
    \centering
    \includegraphics[width=0.47 \textwidth]{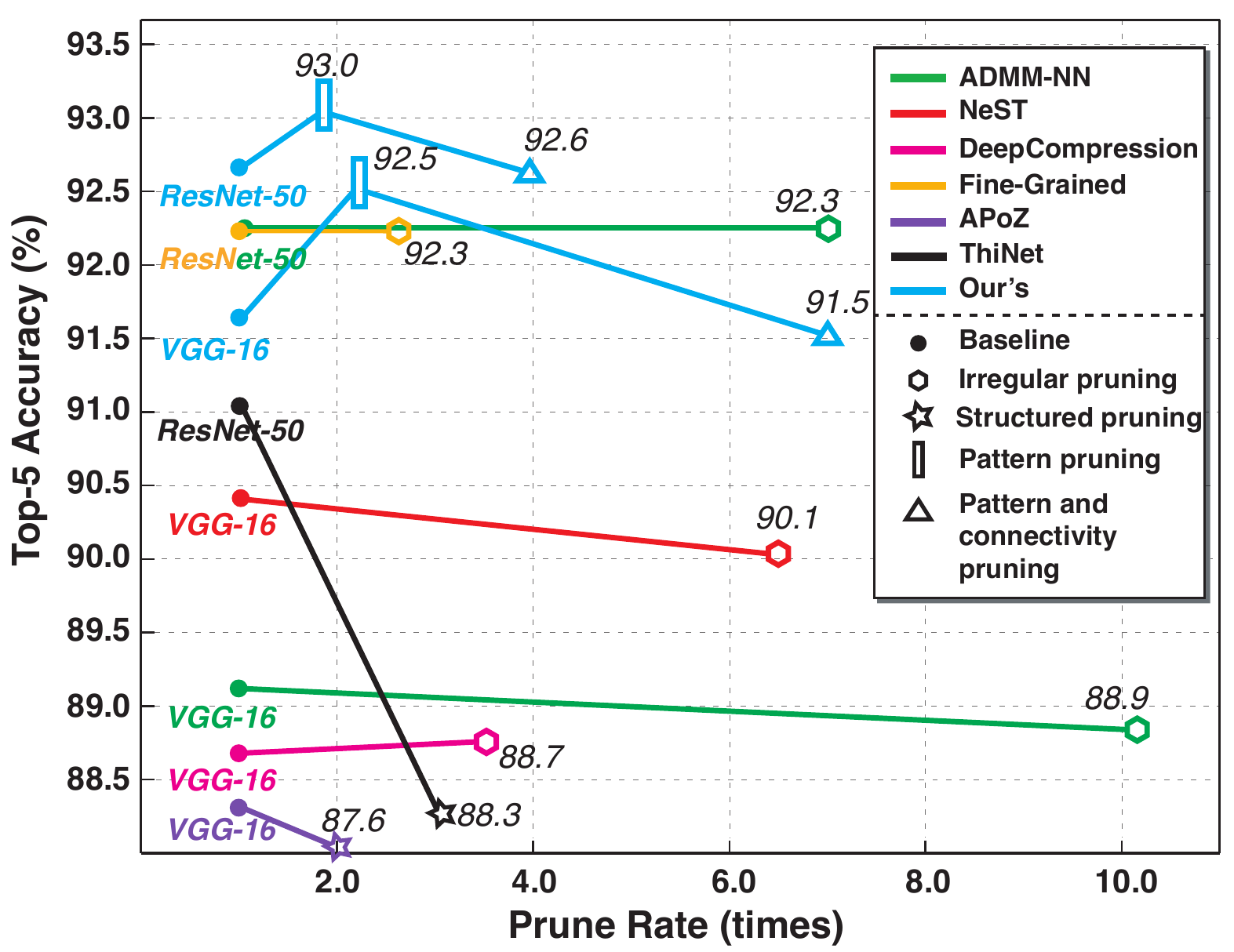}
    %\vspace{-1mm}
    \caption{Comparison results of our pattern and connectivity pruning of VGG-16 and ResNet-50 on ImageNet dataset with: ADMM-NN~\cite{ren2019ADMMNN}, NeST~\cite{dai2017nest}, Deep Compression~\cite{han2015deep}, Fine-grained pruning~\cite{mao2017exploring}, APoZ~\cite{hu2016network} and ThiNet~\cite{luo2017thinet}.}
    \label{fig:acc_result}
    % \vspace{-3mm}
\end{figure}

\subsection{Accuracy Analysis}
In our previous derivation, we have determined the (four) SCPs as our pattern set. Our algorithm-level solution starts from a pre-trained DNN model, or can train from scratch. To generate \projectname model, we need to assign SCPs to each kernel (pattern pruning) or prune specific kernels (connectivity pruning), and train the active (unpruned) weights. To achieve this goal, we extend the ADMM-NN framework in~\cite{ren2019ADMMNN} to produce pattern and connectivity-pruned models. 
% Due to space limit, the algorithm details in \projectname model generation are described in Supplemental Materials.

% \textcolor{red}{our algorithm-level solution}

\textbf{Accuracy results} are illustrated in Figure~\ref{fig:acc_result}. 
% To further validate that our designed SCPs improve accuracy, we set contrast experiments about pruning VGG-16 and ResNet-50 on ImageNet dataset. 
% Figure~\ref{fig:acc_result} indicates that our \emph{Pattern Pruning} could improve accuracy along with model weights pruned. 
% Even started with the state-of-the-art model as baseline, which applies the bag of tricks to reach the highest accuracy, 
Starting from the baseline accuracy results that are in many cases higher than prior work, we have the first conclusion that \emph{the accuracy will improve when applying our designed SCPs on each convolution kernel}.
% Figure~\ref{fig:acc_result} indicates that \emph{pattern pruning} improves accuracy of a relative high baseline model, which is much higher than the baseline model used in prior works.
For ImageNet dataset, \emph{pattern pruning} improves the top-5 accuracy of VGG-16 from $91.7\%$ to $92.5\%$, and ResNet-50 from $92.7\%$ to $93.0\%$ with SCPs applied to each convolution kernel. 
The accuracy improvement is attributed to the enhanced image processing ability of our designed SCPs.
% Combined with \emph{Connectivity Pruning}, 
% more compression rate can be achieved without compromise our baseline accuracy.
% the prune ratio increases further and the accuracy keeps at the state-of-the-art point.

\textbf{Pruning vs. accuracy for non-structured pruning, structured pruning and \projectname.} 
Combined with \emph{connectivity pruning}, 
\projectname achieves higher compression rate without accuracy compromise. Comparing with other pruning methods, i.e., non-structured pruning and structured pruning, we conclude that: (i) \projectname achieves higher accuracy and higher compression rate compared with prior non-structured pruning, and close to the results in ADMM-NN; (ii) compared with structured pruning, under the same compression rate, \projectname achieves higher accuracy, and can structurally prune more weights without hurting accuracy. The detailed comparisons on different sparsity and compression rates are shown in Figure~\ref{fig:acc_result}
% Detailed comparison is shown in Supplemental Materials.

\begin{figure}[t]
    \centering
    \includegraphics[width=0.47 \textwidth]{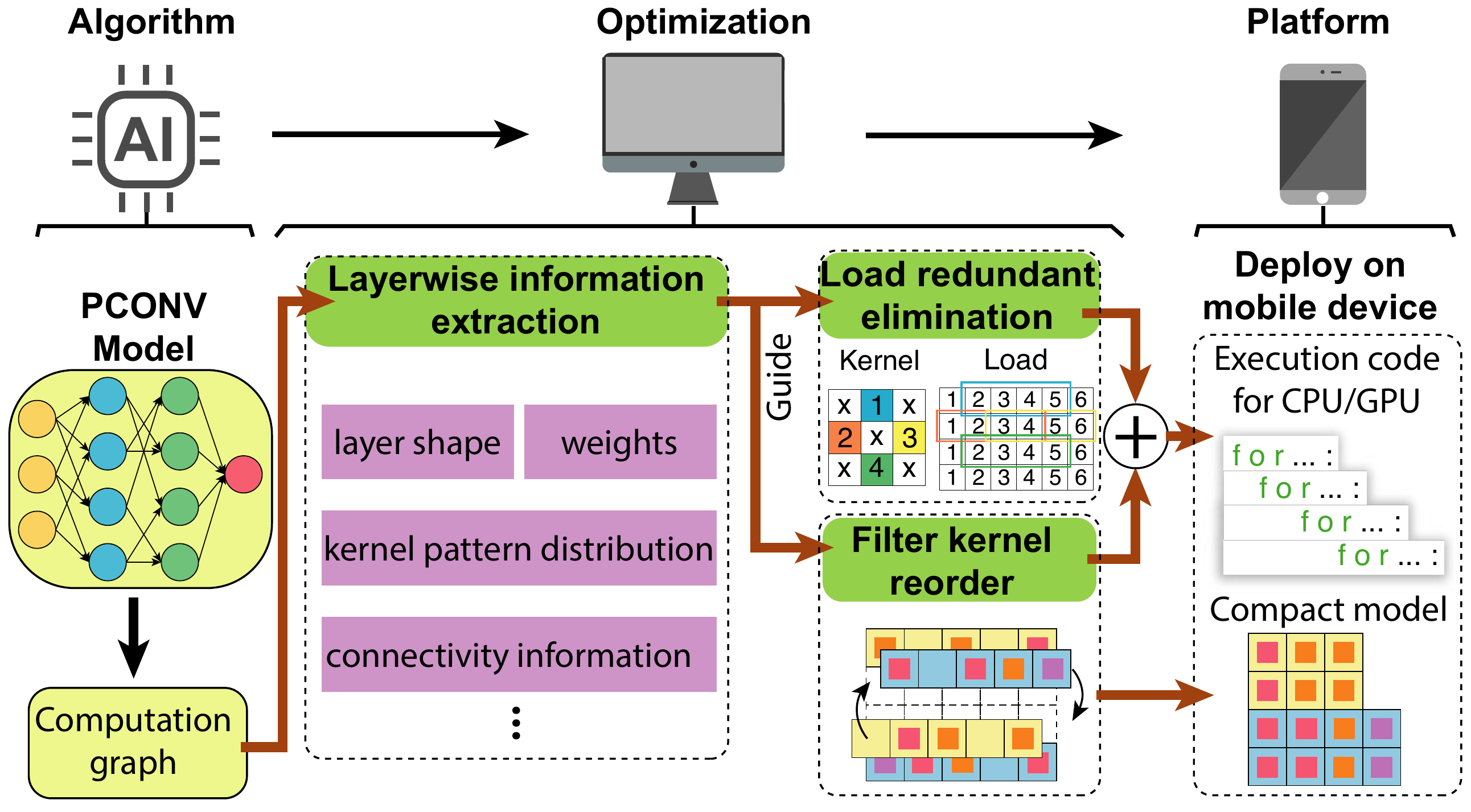}
    %\vspace{-1mm}
    \caption{Overview of \projectname acceleration framework. From algorithm-level design to platform-level implementation.}
    \label{fig:mobi_overview}
    % \vspace{-3mm}
\end{figure}

% \begin{figure}[t]
%     \centering
%     \includegraphics[width=0.4\textwidth]{figs/imagenet_result.pdf}
%     % \atop
%     % \subfigure{
%     %     \begin{minipage}[t]{0.24\textwidth}
%     %     \flushleft
%     %     \includegraphics[scale=0.51]{figs/imagenet_result.pdf}
%     %     \end{minipage}
%     % }
%     % \subfigure{
%     %     \begin{minipage}[t]{0.24\textwidth}
%     %     \centering
%     %     \includegraphics[scale=0.51]{figs/cifar_result.pdf}
%     %     \end{minipage}
%     % }
%     %\vspace{-1mm}
%     \caption{Results of Pattern Pruning and Connectivity Pruning of VGG-16 and ResNet-50 on ImageNet dataset(left) and CIFAR-10 dataset(right)}
%     \label{fig:imagenet_result}
%     % \vspace{-3mm}
% \end{figure}

\section{Compiler-assisted DNN Inference Framework}
In this section, we propose our novel compiler-assisted DNN inference acceleration framework for mobile devices. Motivated by the two merits -- flexibility and regularity of the \projectname model, our compiler-assisted platform uniquely enables \emph{optimized code generation} to guarantee end-to-end execution efficiency. As DNN's computation paradigm is in a manner of layerwise execution, we can convert a DNN model into computational graph, which is embodied by static C++ (for CPU execution) or OpenCL (for GPU execution) code. The code generation process includes three steps as Figure~\ref{fig:mobi_overview} shows: (i) layerwise information extraction; (ii) filter kernel reorder; (iii) load redundancy elimination.

\textbf{Layerwise information extraction} is a model analysis procedure. In particular, it analyzes detailed kernel pattern and connectivity-related information. Key information such as pattern distribution, pattern order and connection between input/output channel through kernels are utilized by the compiler to perform optimizations in steps (ii) and (iii).

\begin{figure}[t]
    \centering
    \includegraphics[width=0.46 \textwidth]{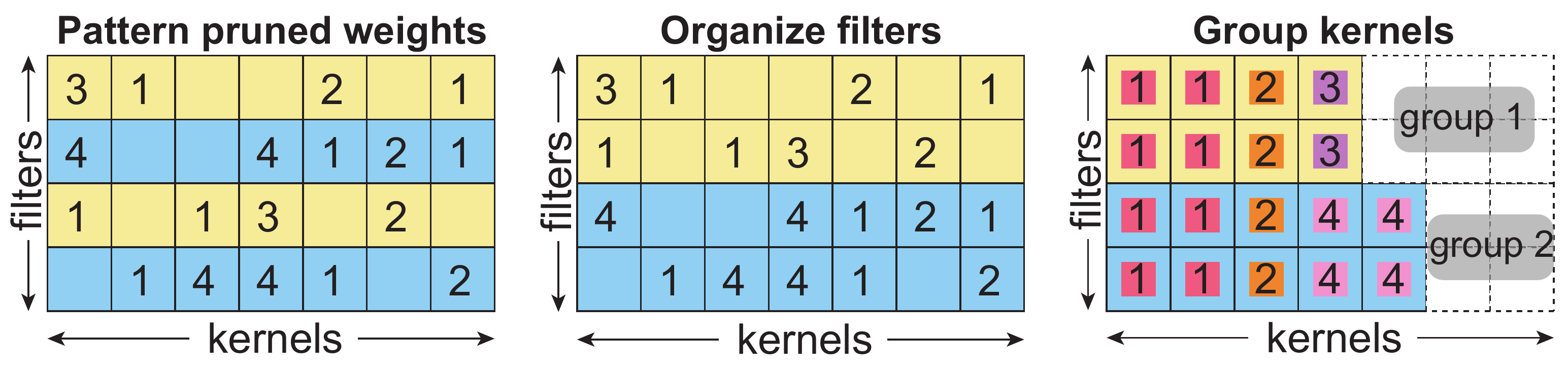}
    %\vspace{-1mm}
    \caption{Steps of filter kernel reorder: each square represents a convolution kernel; the number represents the specific pattern type of this kernel.}
    \label{fig:fkr}
    % \vspace{-3mm}
\end{figure}

\textbf{Filter kernel reorder} is designed to achieve the best of instruction-level and thread-level parallelism. When a \projectname model is trained, patterns and connections of all kernels are already known, i.e., the computation pattern is already fixed before deploying the model for inference. All these information of patterns are collected from layerwise information extraction, and is leveraged by filter kernel reorder to (i) organize the filters with similar kernels together to improve {\em inter-thread} parallelism, and (ii) order the same kernels in a filter together to improve {\em intra-thread} parallelism. Figure~\ref{fig:fkr} illustrates the two key steps of filter kernel reorder: (i) organizes similar filters next to each other; (ii) groups kernels with identical patterns in each filter together. 
As a result, the generated execution code eliminates much of execution branches, implying higher instruction-level parallelism; meanwhile, similar filter groups escalate execution similarity and result in a good load balance, achieving better thread-level parallelism.

% \begin{figure*}[t]
% \flushleft
% \subfigure{
% \flushleft
% \begin{minipage}[t]{0.18\linewidth}
% \centering
% \includegraphics[width=2cm]{figs/snake_original.pdf}
% %\caption{fig1}
% \end{minipage}%

% \begin{minipage}[t]{0.18\linewidth}
% \centering
% \includegraphics[width=2cm]{figs/snake_pytorch_official.pdf}
% %\caption{fig2}
% \end{minipage}%

% \begin{minipage}[t]{0.18\linewidth}
% \centering
% \includegraphics[width=2cm]{figs/snake_pruned.pdf}
% %\caption{fig2}
% \end{minipage}
% }%
% \subfigure{
% }%
% \subfigure{

% }%

% \caption{Visualization intermediate results of VGG-16 baseline and pattern pruned VGG-16}
% \end{figure*}

\textbf{Load redundancy elimination} addresses the issue of irregular memory access that causes memory overhead. In DNN execution, the data access pattern of input/output is decided by the (none-zero elements) patterns of kernels. Therefore, we can generate data access code with this information for each kernel pattern and call them dynamically during DNN execution. Because the data access code consists of all information at kernel-level computation, it is possible to directly access valid input data that is associated with the non-zero elements in a pattern-based kernel. 
% \textcolor{red}{The above sentence. Meaning unclear. Instruct?} 
After steps (i) and (ii), patterns are distributed in a structured manner, which reduces the calling frequency of data access code and as a result, reduces the memory overhead.

% The preserved hardware friendly properties of \emph{PatDNN}, allowing compiler make the best of instruction-level and thread-level parallelism.

\begin{figure*}[t]
    \centering
    \includegraphics[width=0.955 \textwidth]{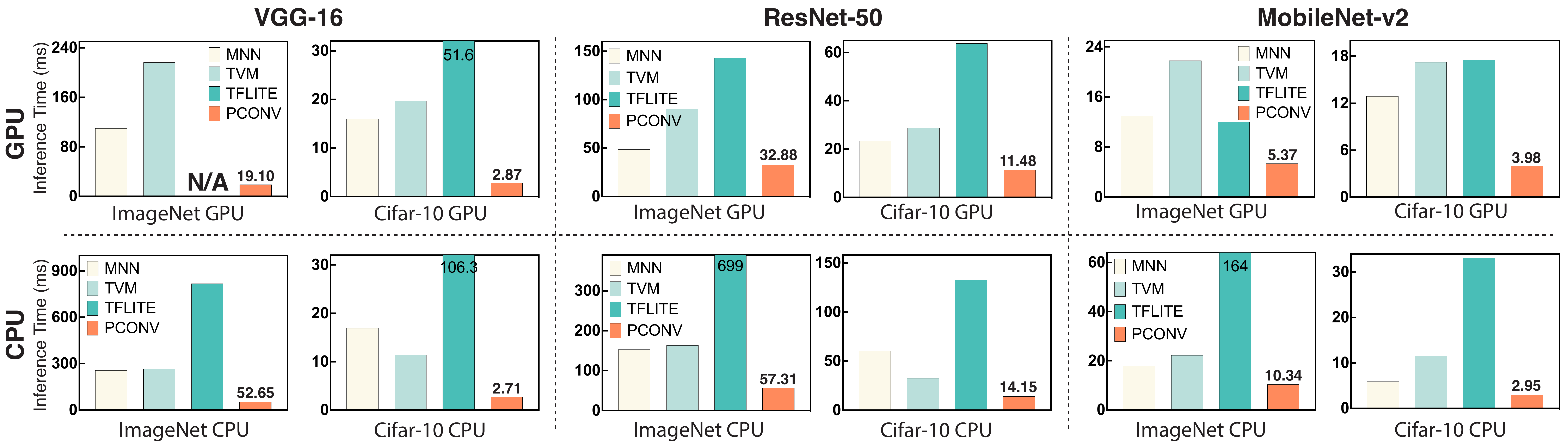}
    %\vspace{-1mm}
    \caption{Mobile CPU/GPU inference time ($ms$) on different network structures inferring Cifar-10 and ImageNet images.}
    \label{fig:speed}
    % \vspace{-3mm}
\end{figure*}

\begin{figure}[b]
    \centering
    \includegraphics[width=0.35 \textwidth]{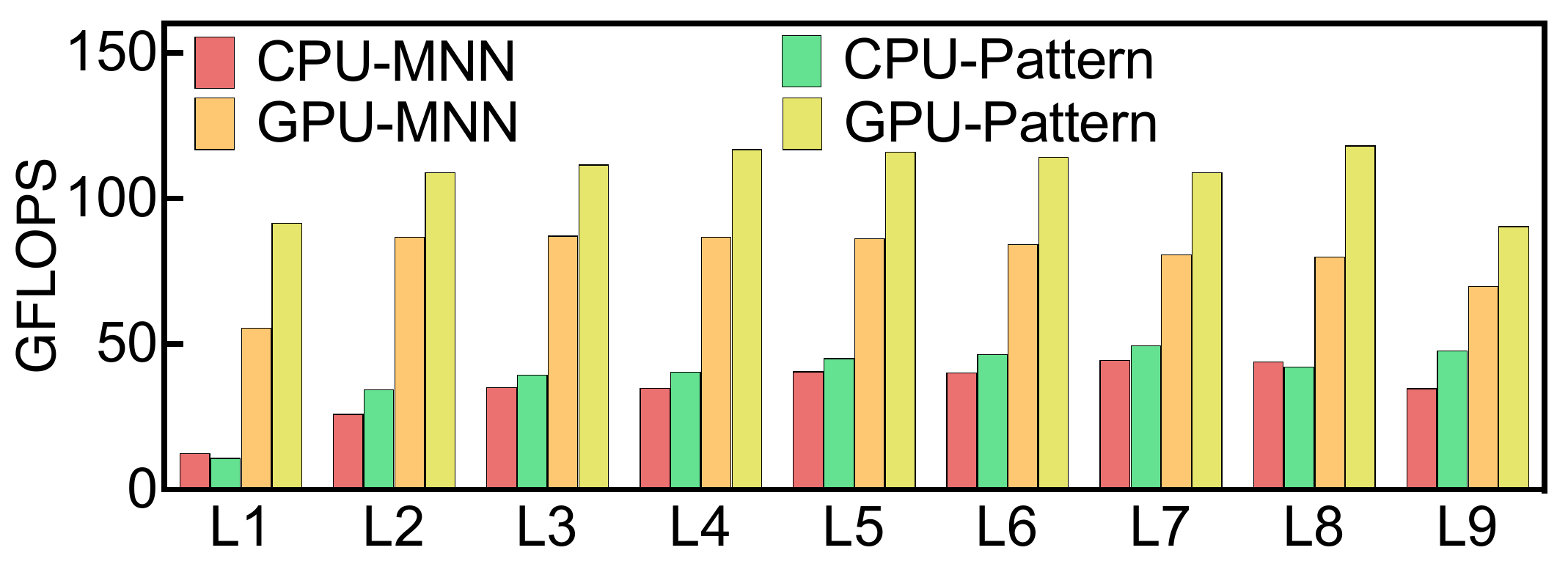}
    %\vspace{-1mm}
    \caption{On-device GFLOPS performance evaluation of MNN and \projectname.}
    \label{fig:gflops}
    % \vspace{-3mm}
\end{figure}

\section{Experimental Results}
In this section, we evaluate the execution performance of our compiler-assisted framework with our \projectname model deployed. 
All of our evaluation models are generated by ADMM pruning algorithm, and are trained on an eight NVIDIA RTX-2080Ti GPUs server using PyTorch.
% To generate \projectname model, we adopt the general ADMM-NN framework in~\cite{ren2019ADMMNN} to produce compressed models. Different from ADMM-NN, which using ADMM~\cite{boyd2011distributed} to prune irregular weights, we use ADMM to assign pruning pattern on each kernel (pattern pruning) or prune a specific kernel (connectivity pruning). The \projectname-oriented ADMM regularization algorithm is proposed in our Supplemental Materials.

\subsection{Methodology}
In order to show acceleration of \projectname on mobile devices, we compare it with three state-of-art DNN inference acceleration frameworks, TFLite~\cite{TensorFlow-Lite}, TVM~\cite{chen2018tvm}, and MNN~\cite{Ali-MNN} using same sparse DNN models. 
Our experiments are conducted on a Samsung Galaxy S10 cell phone with the latest Qualcomm Snapdragon 855 mobile platform that consists of a Qualcomm Kryo 485 Octa-core CPU and a Qualcomm Adreno 640 GPU.

In our experiment, our generated \projectname models are based on three widely used network structures, VGG-16~\cite{simonyan2014very}, ResNet-50~\cite{he2016deep} and MobileNet-v2~\cite{howard2017mobilenets}. 
Since convolution operation is most time-consuming (more than $95\%$ of the total inference time) in DNN computation, our evaluation on the above network structures focus on convolutional layers performance. 
In order to provide a very clear illustration on how \projectname enhances mobile performance, 
% and calculation reduction, 
the whole device-level evaluation is shown in three aspects: (i) execution time, (ii) on-device GFLOPS performance and (iii) how pattern counts affect performance.

\subsection{Performance Evaluation}
In this part, we demonstrate our evaluation results on mobile device from the three aspects we discussed above.
In order to illustrate \projectname has the best acceleration performance on mobile devices, our comparison baselines, i.e., TFLite, TVM and MNN use the fully optimized configurations (e.g., Winograd optimization is turned on).

\textbf{Execution time.} Figure~\ref{fig:speed} shows mobile CPU/GPU performance of \projectname model executing on our compiler-assisted DNN inference framework. 
% To make inference of one image from either Cifar-10 or ImageNet, 
On CPU, \projectname achieves $9.4\times$ to $39.2\times$ speedup over TFLite, $2.2\times$ to $5.1\times$ speedup over TVM and $1.7\times$ to $6.3\times$ speedup over MNN. On GPU, \projectname achieves $2.2\times$ to $18.0\times$ speedup over TFLite, $2.5\times$ to $11.4\times$ speedup over TVM and $1.5\times$ to $5.8\times$ speedup over MNN. 
For the largest DNN (VGG-16) and largest data set (ImageNet), our framework completes computations on a single input image within $19.1ms$ (i.e., 52.4 frames/sec) on GPU,
% One of our result highlights shows that for the network that involves most computations (VGG-16), it only takes $19.1ms$ (52.4 frames/sec) to finish classification for one image from ImageNet, 
which meets the real-time requirement (usually 30 frames/sec, i.e., 33 $ms$/frame).

\textbf{On-device GFLOPS performance.} From the previous comparison results we see that MNN has the higher performance than TVM and TFLite. To show that \projectname has better throughput on mobile devices, we compare \projectname with MNN by measuring their run-time GFLOPS on both CPU and GPU. 
% \textcolor{purple}{To conduct a fair comparison, we turn off the Winograd optimization for MNN.} 
Figure~\ref{fig:gflops} demonstrates layerwise GFLOPS performance comparison between \projectname and MNN. The 9 layers we pick from VGG-16's 13 convolutional layers are representing 9 unique layers with 9 unique layer sizes. 
The other 4 layers are omitted in Figure~\ref{fig:gflops} because they have repeated layer sizes which product repeated GFLOPS results. 
From the results we can see that for both CPU and GPU throughputs, \projectname outperforms MNN. 
% \textcolor{red}{Do we need point out that we didn't use Winograd(The GFLOPS from MNN didn't open Winograd)? }

\textbf{Pattern counts vs. performance.} In order to determine how pattern counts affects execution performance, we design some random patterns with 4 non-zero elements in one kernel alongside with our designed SCPs. Table~\ref{table:patternCount_cifar} and Table~\ref{table:patternCount_imagenet} show accuracy and execution time under different pattern counts using VGG-16 on Cifar-10 and ImageNet datasets. The results show that the accuracy losses are not necessarily related to the increase of pattern counts, but the execution performance drops quickly, especially on ImageNet dataset. 
The pattern counts vs. performance results prove that our designed SCPs result in 
ideal performance with a negligible accuracy loss.

% \textcolor{red}{Are you going to put 6-pattern results here?}\textcolor{blue}{-6-pattern Removed}

% As images from ImageNet are large enough (224$\times$224 pixels in our case) to meet the real-time processing application requirement instead of images from Cifar-10 (32$\times$32 pixels), the pattern counts v.s. performance results provide indirect evidence that proves superiority of our designed SCPs in both DNN performance and platform performance level.

\begin{table}[h]
\scriptsize
    \centering
    % \vspace{-2mm}
    \caption{Pattern counts vs. performance. Evaluation uses model with pattern (2.25$\times$) and connectivity (8.8$\times$) sparsity on VGG-16 Cifar-10 dataset. Top-1 accuracy displayed.}
    \renewcommand{\arraystretch}{0.88}
    \resizebox{\linewidth}{!}{
        \begin{tabular}{|c|c|c|c|c|c|}
            \hline
            \textbf{Dataset} & \textbf{Pattern\#} & \textbf{Acc. (\%)} & \textbf{Acc. loss (\%)} & \textbf{Device} & \textbf{Speed (ms)} \\ \hline
            \multirow{6}{*}{Cifar-10} & \multirow{2}{*}{\textbf{4}} & \multirow{2}{*}{93.8} & \multirow{2}{*}{-0.3} & CPU & 2.7\\ \cline{5-6}
            &  &  &  & GPU & 2.9 \\ \cline{2-6}
            % & \multirow{2}{*}{\textbf{6}} & \multirow{2}{*}{93.5} & \multirow{2}{*}{0} & CPU & 2.7\\ \cline{5-6}
            % &  &  &  & GPU & 2.9 \\ \cline{2-6}
            & \multirow{2}{*}{\textbf{8}}& \multirow{2}{*}{93.7} & \multirow{2}{*}{-0.2} & CPU & 2.9 \\ \cline{5-6}
            &  &  &  & GPU & 3.0 \\ \cline{2-6}
            & \multirow{2}{*}{\textbf{12}}& \multirow{2}{*}{93.8} & \multirow{2}{*}{-0.3} & CPU & 3.1 \\ \cline{5-6}
            &  &  &  & GPU & 3.3 \\ 
            \hline
					 
        \end{tabular}
    }
    \label{table:patternCount_cifar}
\end{table}

\begin{table}[h]
\scriptsize
    \centering
    % \vspace{-5mm}
    \caption{Pattern counts vs. performance. Evaluation uses model with pattern (2.25$\times$) and connectivity (3.1$\times$) sparsity on VGG-16 ImageNet dataset. Top-5 accuracy displayed.}
    \renewcommand{\arraystretch}{0.88}
    \resizebox{\linewidth}{!}{
        \begin{tabular}{|c|c|c|c|c|c|}
            \hline
            \textbf{Dataset} & \textbf{Pattern\#} & \textbf{Acc. (\%)} & \textbf{Acc. loss (\%)} & \textbf{Device} & \textbf{Speed (ms)} \\ \hline
            \multirow{6}{*}{ImageNet} & \multirow{2}{*}{\textbf{4}} & \multirow{2}{*}{91.5} & \multirow{2}{*}{0.2} & CPU & 52.7 \\ \cline{5-6}
            &  &  &  & GPU & 19.1 \\ \cline{2-6}
            % & \multirow{2}{*}{\textbf{6}} & \multirow{2}{*}{91.4} & \multirow{2}{*}{0.3} & CPU & 57.7 \\ \cline{5-6}
            % &  &  &  & GPU & 20.3 \\ \cline{2-6}
            & \multirow{2}{*}{\textbf{8}}& \multirow{2}{*}{91.6} & \multirow{2}{*}{0.1} & CPU & 58.9 \\ \cline{5-6}
            &  &  &  & GPU & 22.0 \\ \cline{2-6}
            & \multirow{2}{*}{\textbf{12}}& \multirow{2}{*}{91.6} & \multirow{2}{*}{0.1} & CPU & 105.2 \\ \cline{5-6}
            &  &  &  & GPU & 32.1 \\ 
            \hline
					 
        \end{tabular}
    }
    \label{table:patternCount_imagenet}
\end{table}

\section{Conclusion}
This paper presents \projectname, a desirable sparsity type in DNN weight pruning that elicits mobile devices acceleration, leading to real-time mobile inference. \projectname inherits the high flexibility in non-structured pruning which helps achieving high accuracy and compression rate, and maintains highly structured weight composition like structured pruning which leads to hardware friendlinesses such as optimized memory access, balanced workload and computation parallelism etc. To show \projectname's real-time performance on mobile devices, we design a compiler-assisted DNN inference framework, which can fully leverage \projectname's structural characteristics and achieve very high inference speed on representative large-scale DNNs.

% \section{Acknowledgement}
% This work is partly supported by the National Science Foundation CCF-1901378, and is partly supported by DiDi GAIA Collaborative Research Funds. 
% We thank all anonymous reviewers for their feedback.

\bibliographystyle{aaai}
\small
% \bibliography{references}

\end{document}